\documentclass[letterpaper,conference,10pt]{IEEEtran}

\usepackage{graphicx}
\usepackage{multirow}
\usepackage{url}
\usepackage{amsmath}
\usepackage{amssymb}

\usepackage[linesnumbered,ruled,vlined]{algorithm2e}

\usepackage{lipsum} 
\usepackage{float} 

\usepackage{hyperref}
\usepackage{dblfloatfix}

\begin{document}

\title{ExFake: Towards an Explainable Fake News Detection Based on Content and Social Context Information}

\author{\IEEEauthorblockN{Sabrine Amri}
\IEEEauthorblockA{DIRO\\
University of Montreal\\
Montreal, Canada\\
sabrine.amri@umontreal.ca}
\and
\IEEEauthorblockN{Henri-Cedric Mputu Boleilanga}
\IEEEauthorblockA{DIRO\\
University of Montreal\\
Montreal, Canada\\
henri-cedric.mputu.boleilanga@umontreal.ca}
\and
\IEEEauthorblockN{Esma Aïmeur}
\IEEEauthorblockA{DIRO\\
University of Montreal\\
Montreal, Canada\\
aimeur@iro.umontreal.ca}}

\maketitle

\begin{abstract}
ExFake is an explainable fake news detection system based on content and context-level information. It is concerned with the veracity analysis of online posts based on their content, social context (i.e., online users' credibility and historical behaviour), and data coming from trusted entities such as fact-checking websites and named entities. Unlike state-of-the-art systems, an Explainable AI (XAI) assistant is also adopted to help online social networks (OSN) users develop good reflexes when faced with any doubted information that spreads on social networks. The trustworthiness of OSN users is also addressed by assigning a credibility score to OSN users, as OSN users are one of the main culprits for spreading fake news. Experimental analysis on a real-world dataset demonstrates that ExFake significantly outperforms other baseline methods for fake news detection.
\end{abstract}


\section{Introduction}
\label{intro}
The world has become a small village where information circulates at high speed. Online social networks (OSN) have allowed anyone to express themselves and share information. Unfortunately, it is not all upsides. There is growing recognition of the adverse effects of OSN. One of the most popular is the spread of mis/dis/mal-information also known as \emph{``fake news''}.
{There is still no unanimous definition of this expression, mainly because its meaning and usage have changed over time\cite{amri2023fake}. It was originally used to refer to false information and the dissemination of sensational information under the pretext of news reporting. The term has now evolved and becomes synonymous with the spread of false information.}

The fake news detection problem has attracted a lot of attention. Various methods for detecting fake news in different forms of data have been proposed~\cite{amri2023fake,olan2022fake}. They can be divided into content and social context-based methods~{\cite{amri2023fake}. Indeed, the content of the news is fully available in the early stages, unlike auxiliary social context-based information (e.g., social engagement, user response, propagation patterns) which can only be obtained after the news has spread. However, detecting fake news from news content alone is not enough as news can be intentionally and strategically created to mimic the truth. Therefore, the exploration of auxiliary information is deemed crucial for the effective detection of fake news. Context-based approaches explore the surrounding data outside of news content, which can be an effective direction and have some advantages in areas where the content approaches based on text classification can run into issues. However, most existing studies implementing contextual methods are limited to the use of sophisticated machine learning techniques for feature extraction and they ignore the usefulness of results coming from techniques such as web search and crowdsourcing, which may save time and help in the early detection and identification of fake content. To make matters worse, there has been little previous work breaking out of the black-box nature of such approaches and focusing on providing explanations to OSN users~\cite{amri2022exmulf}. Such explanations are crucial to reflect news credibility, raise OSN users’ awareness, and ultimately influence their behaviour in protecting the security and privacy of both individuals and society. Studies have shown that it is still difficult for individuals to verify the veracity of a given news content based solely on automatic models, and without further explanation~\cite{przybyla2021classification}. Additionally, humans achieved an average accuracy of 54\% in the task of deception judgment~\cite{bond2006accuracy}. Therefore, identifying fake news must shift to explainable and interpretable automatic detection models to avoid ambiguity and to raise OSN users' awareness. 

Keeping this in mind, and motivated by the lack of powerful tools and the desire to help people, in this paper, we develop a system named \emph{ExFake}, which stands for explainable fake news detection. ExFake is a hybrid system (i.e., relying on both the content and social context of the online post) based on credibility analysis of the source (i.e., OSN user who shared the post) and similarity analysis with data received continuously from a trusted fact-checking organization and extracted named entities (i.e, official Twitter accounts). This system discovers interesting and useful patterns and relationships in large volumes of data that come from a flow of data. This flow of data is received from specific sources and it allows ExFake to integrate the evolution of information over time. Thus, the system does not only process information available when a request (i.e., an input online post) is received, but also processes information that will be available in the near future by continuously retrieving data from trusted fact-checkers and the extracted named entities. This first version of ExFake works only on Twitter. It is built on the composition of four modules, each in charge of particular tasks. Two modules are each responsible for similarity analysis of the content retrieved from fact-checking organizations and named entities respectively. Another module is responsible for the credibility analysis of OSN users who have shared the post. The last module is responsible for the decision-making and explainability tasks. The main contributions of this paper are as follows:
\begin{itemize}
    \item A novel system ExFake leverages content and social context-level information. It processes similarity analysis with data retrieved from trusted and legitimate entities (i.e., fact-checking organizations) and learns useful representations for predicting fake news. Namely, ExFake processes both types of input, online post data and broadcast data received from official accounts of the named entities extracted from the input text (i.e., named entity recognition (NER)) to determine the veracity of the input text. 
    \item A credibility analysis of the OSN user who shared the post (i.e., the source), based on the Bayesian average technique and following the same logic as the five-star rating system for e-commerce sites. The credibility analysis describes how likely a user is to post false information based on their historical sharing behaviour.
    \item A neural network-based decision-making module to compute the final percentage of confidence of the input post. 
    \item An explanation module based on Explainable AI (XAI) and Natural Language Inference (NLI) techniques, which is a new variation of existing explanation methods.
    \item An implementation and evaluation using the publicly available FakeNewsNet dataset, enhanced with the features we add (i.e., the legitimacy score and posts and data retrieved from named entities).
\end{itemize}

To our knowledge, this work is the first approach based on the analysis of the content, social context and external evidence such as trusted sources and named entities together with explainability tasks (i.e., XAI). The encapsulation, which consists of finding the closest text (i.e., articles or text) and then determining the inference from the closest retrieved texts in terms of text similarity is a type of combination, which is also used for the first time for the fake news detection problem. 
The rest of this paper is structured in the following manner: In Section~\ref{sc:relatedWk} we provide a review of related work regarding the fake news detection problem from content, context-based, and explainable AI (XAI) perspectives. The system description is presented in Section~\ref{sc:approach}. The analysis and evaluation are provided in Section~\ref{sc:analEval}. Finally, conclusions and future directions are drawn in Section~\ref{sc:conc}.

\section{Related Work}\label{sc:relatedWk}
The present study is built on three existing research axes. First, on methods to detect fake news based on content analysis (i.e.,~related to the content of the news post). Second, on methods based on the analysis of the social context (i.e.,~ associated with the social context of the news post) for the detection of fake news. Third, on expanding fake news classification models using Explainable AI (XAI) to help OSN users understand how a certain classification result was achieved. In our system, all aspects are taken into account (i.e., content and context-based fake news detection as well as explainability).
\subsection{Content and Context-based Fake News Detection Approaches}
The content-based detection approaches are mainly based on data extracted from the content of the input. They rely on content-based detection cues (i.e.,~text and multimedia-based cues), which are features extracted from the content of the news post~\cite{amri2023fake}. Text-based cues are features extracted from the text of the news~\cite{elhadad2019novel}, whereas multimedia-based cues are features extracted from the images and videos attached to the news~\cite{demuyakor2022fake}. Recent directions tend to do a mixture by using multimodal aspects~\cite{amri2022exmulf,chen2022cross} that rely on analyzing both textual and visual data extracted from the news content.
Unlike news content-based solutions, the social context-based approaches capture the skeptical social context of the online news (i.e., the surrounding data) rather than focusing on the news content. Social context-based aspects can be classified into two subcategories~\cite{amri2023fake}, user-based and network-based. User-based aspects refer to information captured from OSN users such as user profile information~\cite{shu2019role} and user behaviour~\cite{MOYCCCJTH} such as user engagement~\cite{uppada2022novel} and response~\cite{zhang2019reply}. Meanwhile, network-based aspects refer to information captured from the properties of the social network where the fake content is shared and disseminated such as news propagation path (e.g.,~propagation time and temporal characteristics of propagation)~\cite{liu2018early}, diffusion patterns (e.g.,~number of retweets, shares)~\cite{shu2019beyond}, as well as user relationships (e.g.,~friendship status among users)~\cite{mishra2020fake}.
Recent directions tend to do a mixture by using both news content and social context-based approaches for fake news detection~\cite{shu2020fakenewsnet}. We believe that the most important social context aspect of fake news detection is the user-based aspect, as the user is the major culprit in creating and spreading fake news. 

\subsection{Explainable AI Approaches}
Although the progress in the detection of fake news has been significant, limited effort has been devoted to explainability. Explainable AI (XAI) systems have recently become a promising direction to achieve transparency in many applications such as fake news detection in online social networks. XAI approaches focus on providing explanations to online social network (OSN) users. 
Therefore, researchers in this category are using multiple techniques and models to incorporate explainability in their prediction models for fake news detection. These techniques include, but are not restricted to, neural network~\cite{shu2019defend}, co-Attention network~\cite{lu2020gcan}, 
natural language processing (NLP)~\cite{denaux2020linked}, network embedding learning~\cite{silva2021propagation2vec}, Tsetlin machine (TM)~\cite{bhattarai2021explainable} and local interpretability (i.e., Local Interpretable Model-Agnostic Explanations (LIME))~\cite{amri2022exmulf}.

\section{The Proposed ExFake System}\label{sc:approach}
{In this section, we give the details of the proposed system, which we call ExFake (Explainable Fake news detection). It is based on continuous data collection and a continuous flow of data coming from specific sources. The system consists of two main parts. The first part is for data processing and source assessment, and it includes three modules (i.e., Ex-Fact, Ex-Source, and Ex-Entity). The second part contains a single module (i.e., Ex-Decision) responsible for calculating the confidence percentage of the input data while providing an explanation on this basis. Specifically, ExFake is composed of four modules, whose architecture is shown in Figure~\ref{fig:DS_Exfake}.} The pseudo-code shown in Algorithm~\ref{alg:Ex-Fake} presents the execution of Ex-Fake once a request (i.e., an online post) is received. The input of Ex-Fake is a post on Twitter. The first step is the extraction and preprocessing of data (i.e., of the published date and the published time of the input tweet).
\begin{figure}[h!]
    \centering    
    \includegraphics[width=\columnwidth]{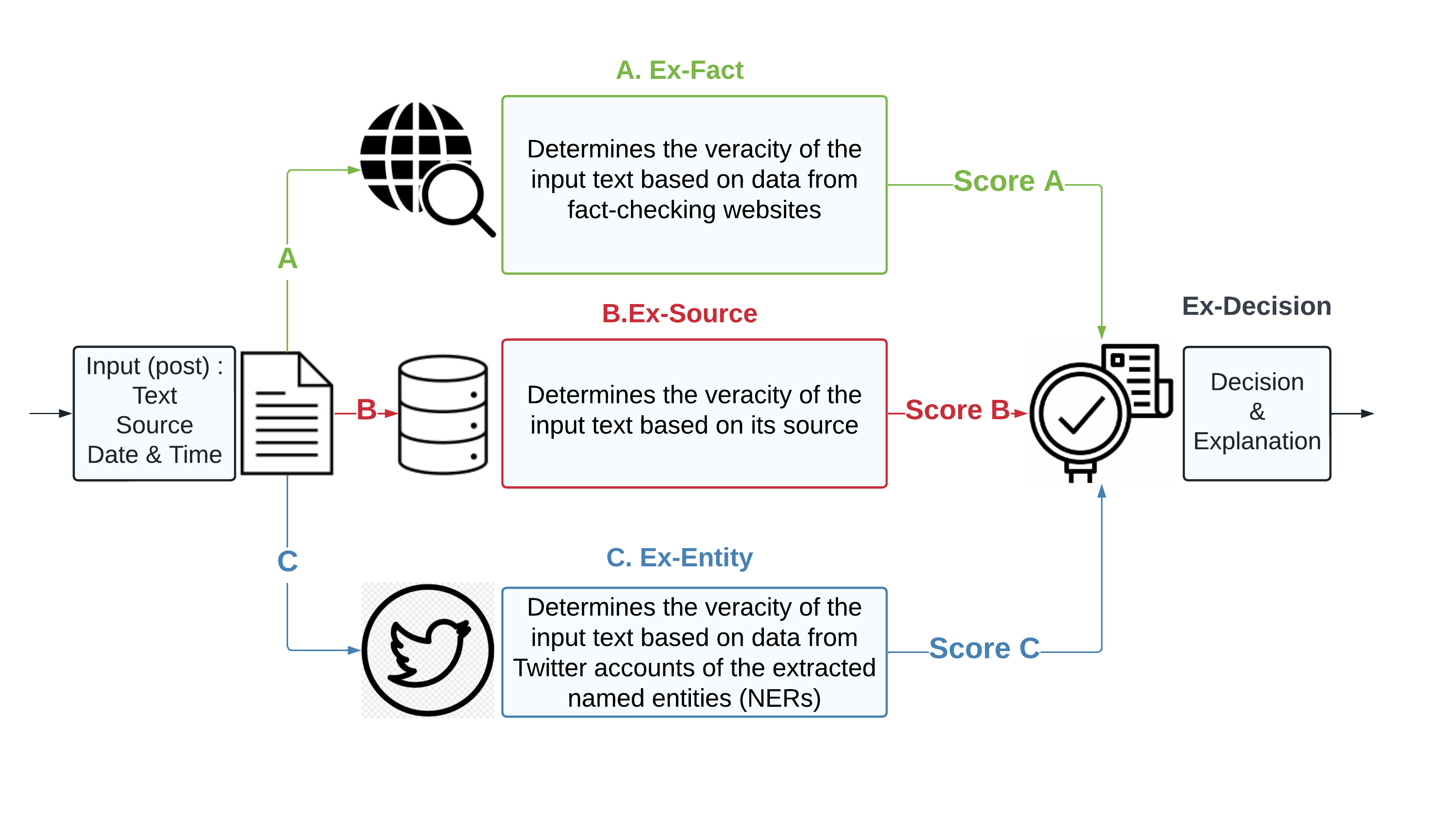}
    \caption{ExFake Architecture}
    \label{fig:DS_Exfake}
\end{figure}

\begin{algorithm}[h]
    \DontPrintSemicolon 
    \SetKwBlock{DoParallel}{do in parallel}{end}
    \KwIn{online post: text, source, date and time}
    \KwOut{percentage of confidence and explanation}
    \emph{data preprocessing()}\;
    \For{$i\leftarrow 0$ \KwTo $2$}{

        \DoParallel{
            Ex-Fact() \;
            Ex-Source() \;
            Ex-Entity() \;
        }
        \emph{sleep(t)} \tcp*[h]{freezes the execution of the loop.}\;
        \emph{Ex-Decision(t)}  \tcp*[h]{computes the final percentage and generates the explanation at the timestamp t.}\;
    }
    \emph{update()} \tcp*[h]{updates the dataset and the legitimacy score of all users}\;
    \Return{Decision: (percentage, explanation)}\;
    \caption{Ex-Fake Algorithm}\label{alg:Ex-Fake}
\end{algorithm}

This system returns three percentages at three different time steps as represented by the for-loop. Even though a percentage is returned in each time step, ExFake keeps receiving a flow of data (i.e., from fact-checking websites and accounts of named entities extracted from the input text) during the freezing time due to the sleep() function. Once all the scores are computed, the module Ex-Decision computes the final percentage of confidence of the input post and generates the explanation. Lastly, the dataset and the legitimacy score of all users are updated and the final decision is made.

\subsection{Ex-Fact}
This module, as illustrated in Figure~\ref{fig:DS_fact}, is in charge of computing a score based on metadata extracted from the input post (i.e., text, date and time) and data coming from trusted fact-checking websites (i.e., includes only PolitiFact.com due to the limitation of our dataset). The first task of this module is then to find articles that are closely related to the input post among all past and upcoming articles of the fact-checking website. With the most similar articles, this module computes a score based on the inference relationship with the input post. 

Among all articles that meet this threshold, we employ the following rules to calculate the final score. 

The output score of the text similarity task ranges from -1 to 1, which we then normalize to a range of 0 to 1. If the normalized score of an article is equal to or greater than the threshold value of 0.8, it is deemed similar to the input post. An article is considered to be close to the input post if the normalized score is above or equal to the threshold of 0.8. Among all articles that meet this threshold, we employ the following rules to calculate the final score:
\begin{itemize}
\item An article that entails the input post is worth 100 points.
\item A neutral article is worth 50 points.
\item An article that contradicts the input post is worth 0 points.
\end{itemize}

Ex-Fact returns the average of the obtained points as the final score.

\begin{figure}[h!]
    \centering
    \includegraphics[width=\columnwidth]{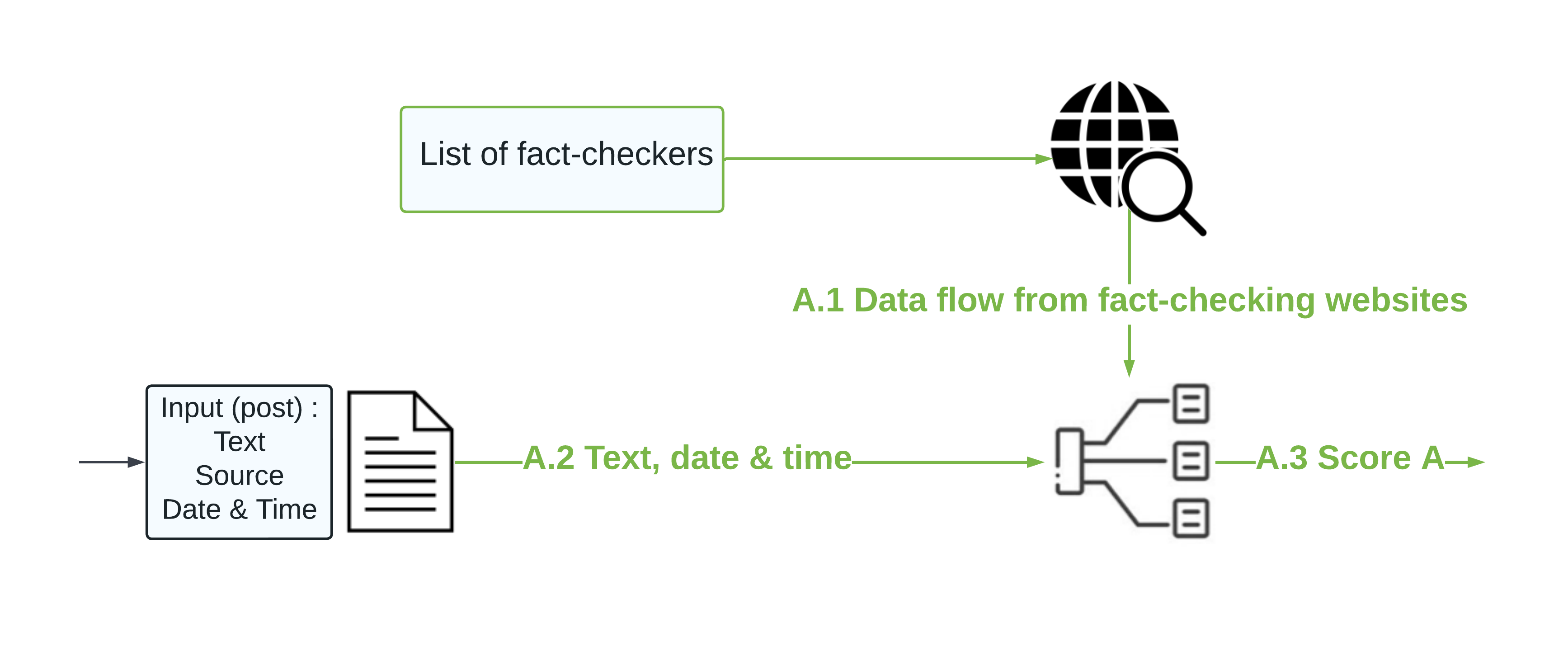}
    \caption{Ex-Fact Module}
    \label{fig:DS_fact}
\end{figure}

\subsection{Ex-Source}
This module, as illustrated in Figure~\ref{fig:DS_source}, computes and returns the legitimacy score (i.e., credibility score) of the online user (i.e., source) who posted the input post based on their past online posts (i.e., posting history). The credibility score considers the history of the percentages of credibility of each post of all the sources to determine the score of each source by using the Bayesian average. Thus, Ex-Source retains the previous percentages of each user's past posts. After each complete execution of the system following a request (i.e., an input post), the score of all users is updated. The more trustworthy posts a user posts, the higher their legitimacy score. When dealing with a new user, this module will give them a score of 50\%.

\begin{figure}[h!]
    \centering
    \includegraphics[width=\columnwidth]{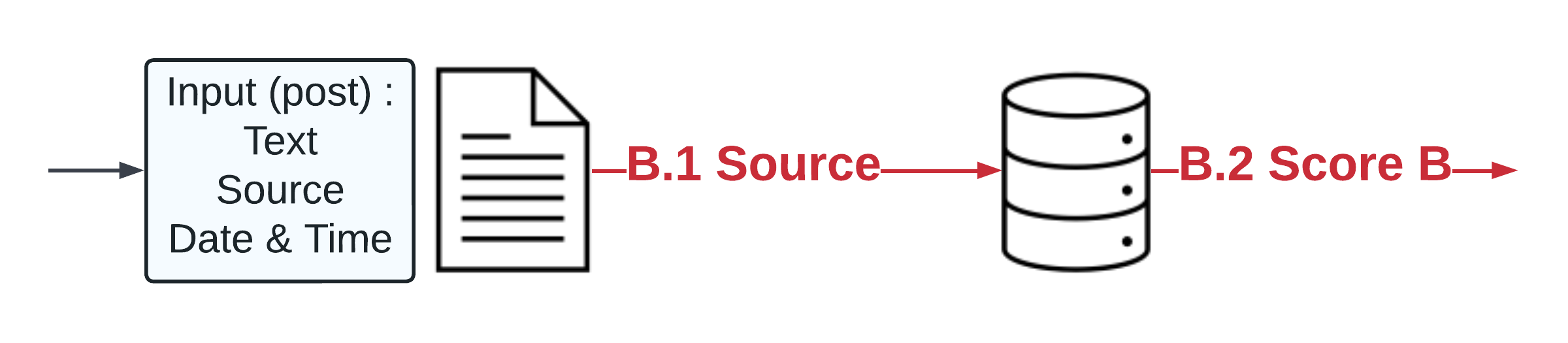}
    \caption{Ex-Source Module}
    \label{fig:DS_source}
\end{figure}

The idea behind the legitimacy score follows the same logic as the five-star rating system for e-commerce sites. The technique used for the legitimacy score is the Bayesian average\footnote{\,\url{https://www.algolia.com/doc/guides/managing-results/must-do/custom-ranking/how-to/bayesian-average/}, last access date: 17-2-2023.}. It ensures that users with lower numbers of received posts (i.e., below a threshold) have less weight in the ranking. For instance, a user who has over one hundred tweets in history that are all true is more credible than a second user with two tweets that are all true as well. His legitimacy score will therefore be higher than that of the second user since the system treated more trustworthy posts from him. The legitimacy score $(LS)$ of a user/source as defined in Eq.~\eqref{eq_LS} returns a value between 0 and 100. The higher this score is, the more posts from this user/source can be trusted. 

\begin{equation}
\label{eq_LS}
     LS = \frac{M \times N + D \times E}{N + E}
\end{equation}

Where $M$ is the mean of the percentage of all the received posts from this user/source, $N$ is the number of posts received from this user/source, $D$ is the mean of all percentages across the whole database and $E$ is the minimum number of received posts to be listed. We set $E$ to 1. 

We chose to use the post history instead of assigning a score to a user based on features such as his localization, job title, number of followers, etc. We did so, primarily because of rare research that used the post history on Twitter. And when it is used, a score is given to a user simply based on the resulting average of his previous tweets. However, unlike these researchers, the credibility score in ExFake (named the legitimacy score) assigns a score to a user based on their previous tweets and the previous tweets of all users using the Bayesian average. 
 Experimentation on this module has shown that the legitimacy score, when used alone, performs better than most state-of-the-art baselines.

\subsection{Ex-Entity}
As illustrated in Figure~\ref{fig:DS_entity}, this module works similarly to the Ex-Fact module. The main difference is the source of the external data. Ex-Entity receives data from legitimate entities, which are the named entities mentioned in the input post. The first step of this module is then to extract all the named entities from the input post (i.e., named entity recognition (NER)). The second step is to retrieve data from those named entities. Then, Ex-Entity finds similar posts posted by the accounts of the extracted named entities and computes a score based on the inference relationship with the input.
\begin{figure}[h!]
    \centering
    \includegraphics[width=\columnwidth]{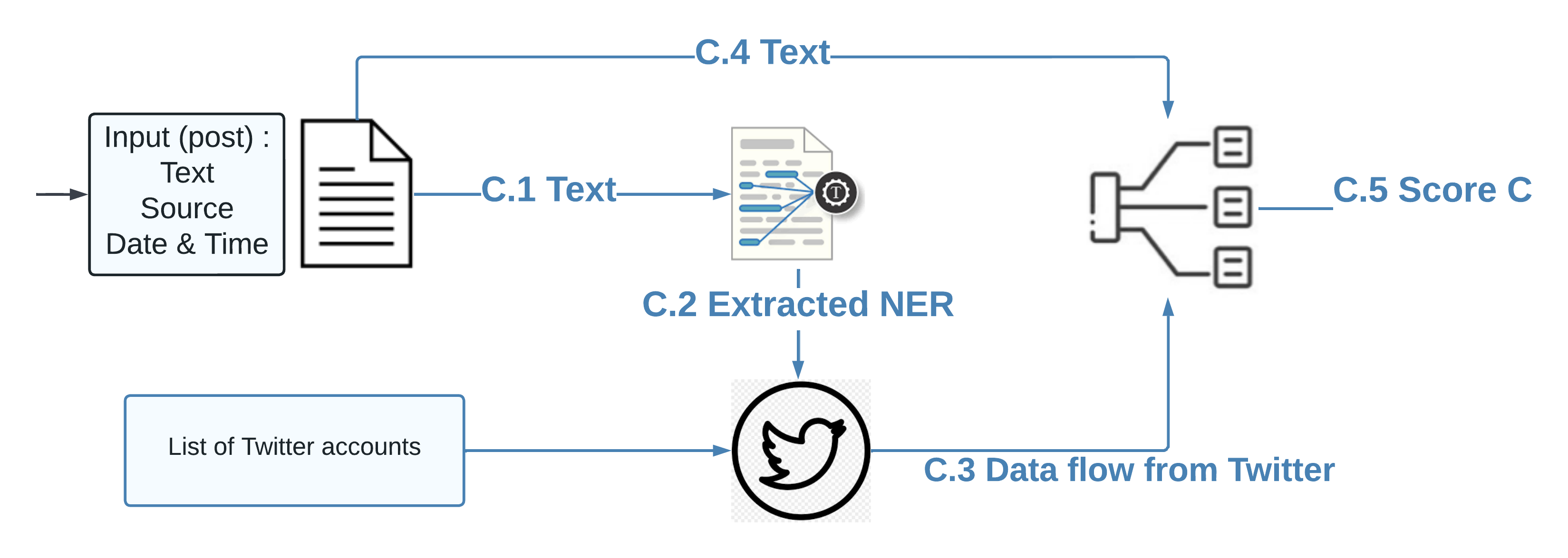}
    \caption{Ex-Entity Module}
    \label{fig:DS_entity}
\end{figure}

For instance, given the following input tweet example "Clinton: Trump called women dogs: At the first presidential debate at Hofstra University, Hillary… \#Skibabs", the module finds the following named entities: Trump, Hofstra University, and Hillary. ExFake has a database of the official certified Twitter accounts of the most mentioned named entities. ExEntity starts to retrieve flows of data from the official accounts of Donald Trump and Hillary Clinton. Then it finds similar tweets from named entities that infer the input tweet.  If the input post does not have any named entity or if we could not retrieve any relevant named entity that meets the threshold, the module will give a score of 50\%.

\subsection{Ex-Decision}
This module, as described in Figure \ref{fig:DS_explain}, is in charge of decision making and it has two tasks to do. The first task is to compute the final percentage of confidence of the input post based on the scores received from the three previous modules (i.e., Ex-Fact, Ex-Source and Ex-entity). The second task is to provide an explanation to the users of the system. The explanation can come from two of the previous modules (i.e., Ex-Fact and Ex-Entity). For the Ex-Fact and Ex-Entity modules, an explanation is returned based on the last NLP task of each of these modules, which is natural language inference (NLI). With the explanation given, OSN users can understand the background of the percentage of confidence their input posts received and thus begin to check their trustworthiness before sharing them.  

\begin{figure}[h!]
    \centering
    \includegraphics[width=\columnwidth]{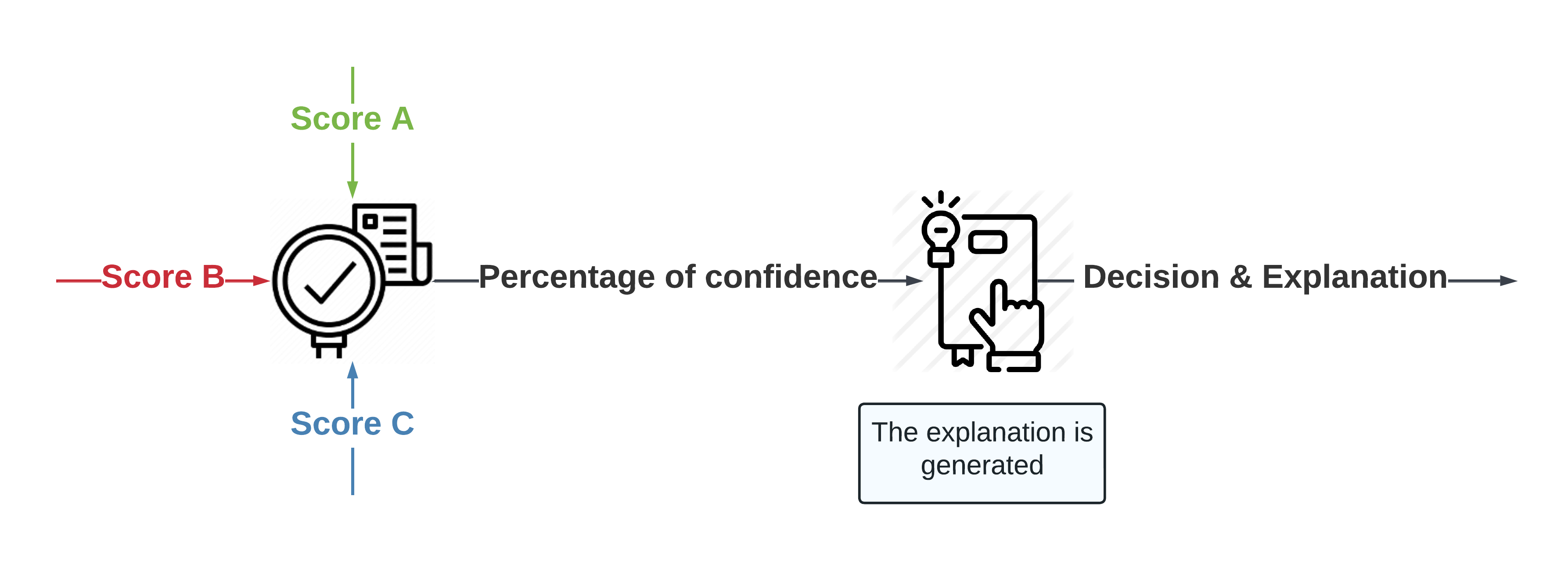}
    \caption{Ex-Decision Module}
    \label{fig:DS_explain}
\end{figure}

Ex-Decision computes the percentage of confidence with a neural network model. Figure \ref{fig:nn} shows the architecture of the used neural network. The score from the modules Ex-Fact (score A), Ex-Source (score B), and Ex-Entity (score C) are the inputs $X_i$. The inputs are multiplied by the neuron weights $W_i$.  After that, a bias $b$ is added to those products. The sum $Z$ of the result of the calculation of each input is then passed to a sigmoid activation function $f(z)$ as defined in Eq.~\eqref{eq_fz}. The result of the activation function is a value between 0 and 1. It is then transformed into a percentage, which is the final percentage of confidence of the input post (i.e., tweet).

\begin{equation}
\label{eq_fz}
f(z) = \frac{1}{1+e^{-x}} 
\end{equation}

\begin{figure}[h!]
    \centering
    \includegraphics[width=\columnwidth]{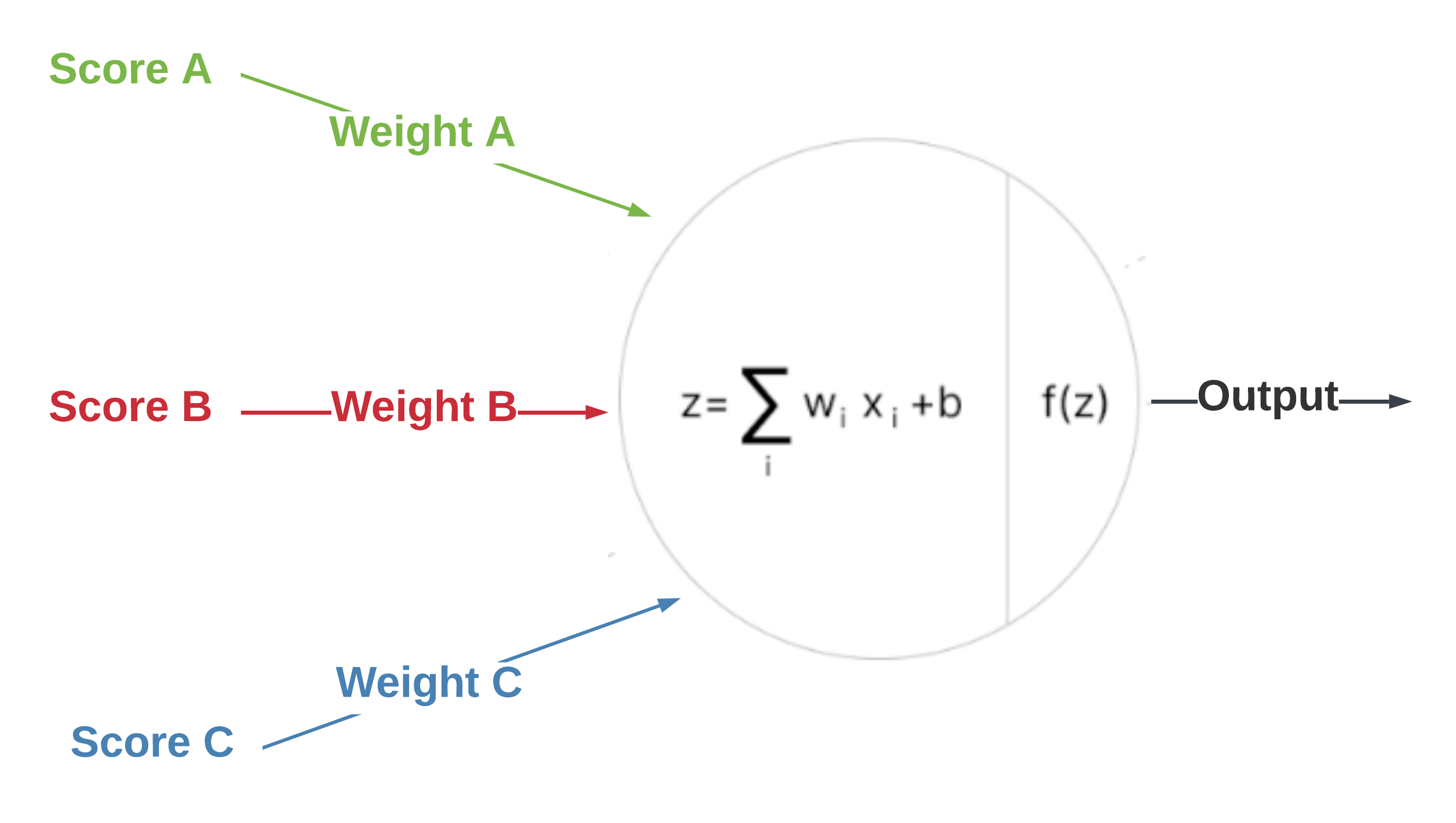}
    \caption{Ex-Decision Neural Networks}
    \label{fig:nn}
\end{figure} 

Unlike most systems that include an explainer which finds and highlights important words within the input text, {we wanted to have an explainer that can help users of our system to develop a particular behaviour when faced with information on social networks.} Thus, Ex-Decision returns the text from trusted organizations and legitimate entities and highlights words that most influenced the result obtained within these trusted and legitimate texts. Trusted organization text is an article from a fact-checking website, and legitimate entity text is a post from a named entity mentioned in the input text. This module also returns the information about the sources who published the post, the {published dates and times} and links to the articles or the tweets. The purpose is to demonstrate how simple it is to confirm certain information by visiting the mentioned named entity’s official account or a reliable website.

\section{Analysis and Evaluation}\label{sc:analEval}
This section describes the experimental evaluation of our method compared to benchmark methods on the FakeNewsNet data repository~\cite{shu2020fakenewsnet}. We first start with detailing the evaluation metrics, the benchmark used in the comparative analysis, and the data collection and processing. This is followed by our empirical results on the FakeNewsNet dataset, as well as a discussion of the results obtained.

\subsection{Evaluation Metrics}
ExFake is a multi-class classification problem. Unlike binary-class classification problems, ExFake returns a percentage of confidence. The percentage of confidence tells how likely it is that we can trust the content of the input tweet and is mapped following the rating labels defined in the PolitiFact dataset as shown in Table~\ref{tab:maping}. Therefore, to evaluate the performance of ExFake, we used the macro-accuracy, macro-precision, macro-recall and macro-F1-score metrics as defined in Table~\ref{tab:metrics}.  

\begin{table}[htpb]
\caption{Mapping scheme of the PolitiFact dataset labels to the percentage of confidence}\label{tab:maping}
\centering
\begin{tabular}{|l|l|}
\hline
\textit{\textbf{Dataset label}} & \textit{\textbf{ExFake percentage of confidence}} \\ \hline
True & {[}87\% - 100\%{]} \\ \hline
Mostly true & {[}70\% - 86\%{]} \\ \hline
Half true & {[}53\% - 69\%{]} \\ \hline
Barely true & {[}37\% - 52\%{]} \\ \hline
False & {[}21\% - 36\%{]} \\ \hline
Pants on fire & {[}0\% - 20\%{]} \\ \hline
\end{tabular}
\end{table}

\begin{table}[htpb]
\caption{Definition of classification evaluation metrics used in this study}\label{tab:metrics}
\centering
\begin{tabular}{|l|l|}
\hline
 Macro-accuracy & $Macro_{accuracy} = \frac{\sum\limits_{k=1}^KAccuracy_k}{K}$ \\ \hline
 $Accuracy_k$ & \begin{tabular}[c]{@{}l@{}}  $A_k = \frac{TP+TN}{TP+TN+FP+FN}$  \\ \end{tabular} \\ \hline
 Macro-precision & $Macro_{precision} = \frac{\sum\limits_{k=1}^KPrecicion_k}{K}$  \\ \hline
 $Precision_k$ & \begin{tabular}[c]{@{}l@{}} $P_k = \frac{TP}{TP+FP} $ \\ \end{tabular} \\ \hline
 Macro-recall & $Macro_{recall}  = \frac{\sum\limits_{k=1}^KRecall_k}{K} $  \\ \hline
 $recall_k$ & \begin{tabular}[c]{@{}l@{}}  $R_k = \frac{TP}{TP+FN}$ \\ \end{tabular} \\ \hline
 Macro-F1-score  & \begin{tabular}[c]{@{}l@{}} $Macro_{F_1} = \frac{\sum\limits_{k=1}^K F_{1_k}}{K}$  \\ \end{tabular} \\ \hline
 $F_1 Score_k$ & \begin{tabular}[c]{@{}l@{}} $F_{1_k} = 2\times \frac{P_k \times R_k}{P_k+R_k} $ \\ \end{tabular} \\ \hline
\end{tabular}
\end{table}

Where $K$ is the number of labelling classes, $TP$ refers to $True Positive$ and represents the set of detected fake news, which are actually fake news, $TN$ refers to $True Negative$ and represents the number of negative samples that have been correctly labelled as negative, $FP$ refers to $False Positive$ and represents the set of detected fake news, which are actually real news, and $FN$ refers to $False Negative$ and represents the set of detected real news, which are actually fake news.
\begin{table*}[!ht]
\caption{Statistics of the FakeNewsNet repository}\label{tab:FakeNewsNet}
\centering
\begin{tabular}{|l|l|l|ll|}
\hline
\multirow{2}{*}{\textit{\textbf{}}} & \multirow{2}{*}{\textit{\textbf{Category}}} & \multirow{2}{*}{\textit{\textbf{Features}}} & \multicolumn{2}{l|}{\textit{\textbf{PolitiFact}}} \\ \cline{4-5} 
 &  &  & \multicolumn{1}{l|}{\textit{\textbf{Fake}}} & \textit{\textbf{Real}} \\ \hline
\multirow{2}{*}{\textit{\textbf{NewsContent}}} & Linguistic & \begin{tabular}[c]{@{}l@{}}\# News articles \\ \# News articles with text\end{tabular} & \multicolumn{1}{l|}{\begin{tabular}[c]{@{}l@{}}432\\ 420\end{tabular}} & \begin{tabular}[c]{@{}l@{}}624\\ 528\end{tabular} \\ \cline{2-5} 
 & Visual & \# News articles with images & \multicolumn{1}{l|}{336} & 447 \\ \hline
\multirow{4}{*}{\textit{\textbf{SocialContext}}} & User & \begin{tabular}[c]{@{}l@{}}\# Users posting tweets \\ \# Users involved in likes \\ \# Users involved in retweets \\ \# Users involved in replies\end{tabular} & \multicolumn{1}{l|}{\begin{tabular}[c]{@{}l@{}}95,553\\ 113,473\\ 106,195\\ 40,585\end{tabular}} & \begin{tabular}[c]{@{}l@{}}249,887\\ 401,363\\ 346,459\\ 18,6675\end{tabular} \\ \cline{2-5} 
 & Post & \# Tweets posting news & \multicolumn{1}{l|}{164,892} & 399,237 \\ \cline{2-5} 
 & Response & \begin{tabular}[c]{@{}l@{}}\# Tweets with replies \\ \# Tweets with likes \\ \# Tweets with retweets\end{tabular} & \multicolumn{1}{l|}{\begin{tabular}[c]{@{}l@{}}11,975\\ 31,692\\ 23,489\end{tabular}} & \begin{tabular}[c]{@{}l@{}}41,852\\ 93,839\\ 67,035\end{tabular} \\ \cline{2-5} 
 & Network & \begin{tabular}[c]{@{}l@{}}\# Followers \\ \# Followees Average \\ \# followers Average \\ \# followees\end{tabular} & \multicolumn{1}{l|}{\begin{tabular}[c]{@{}l@{}}405,509,460\\ 449,463,557 \\ 1299.98 \\ 1440.89\end{tabular}} & \begin{tabular}[c]{@{}l@{}}1,012,218,640 \\ 1,071,492,603 \\ 982.67 \\ 1040.21\end{tabular} \\ \hline
\multirow{2}{*}{\textit{\textbf{\begin{tabular}[c]{@{}l@{}}Spatiotemporal\\ Information\end{tabular}}}} & Spatial & \begin{tabular}[c]{@{}l@{}}\# User profiles with locations \\ \# Tweets with locations\end{tabular} & \multicolumn{1}{l|}{\begin{tabular}[c]{@{}l@{}}217,379\\ 3,337\end{tabular}} & \begin{tabular}[c]{@{}l@{}}719,331\\ 12,692\end{tabular} \\ \cline{2-5} 
 & Temporal & \begin{tabular}[c]{@{}l@{}}\# Timestamps for news pieces \\ \# Timestamps for response\end{tabular} & \multicolumn{1}{l|}{\begin{tabular}[c]{@{}l@{}}296\\ 171,301\end{tabular}} & \begin{tabular}[c]{@{}l@{}}167\\ 669,641\end{tabular} \\ \hline
\end{tabular}
\end{table*}
\subsection{Benchmark Models}
We used seven learning models, used by Shu et al.\cite{shu2020fakenewsnet} in the FakeNewsNet, in conjunction with our proposed ExFake system to compare and evaluate its performance for the fake news detection task. The seven models include four conventional machine learning models and three versions of Social Article Fusion (SAF) models.
The conventional machine learning models include Support Vector Machines (SVM), Logistic Regression (LR), Naive Bayes (NB), and Convolutional Neural Network (CNN). The Social Article Fusion (SAF) models include SAF/S (i.e., relies on the news content), SAF/A (i.e.,  relies on the social context: the temporal pattern of user engagements), and SAF (i.e., a combination of SAF/S and SAF/A.).

\subsection{Data Collection and Processing}
\subsubsection{Dataset}
The lack of comprehensive and community-driven fake news datasets is one of the main issues that researchers face. Not only are existing datasets sparse, but they also lack a variety of aspects commonly needed in research, such as news content, social context, and spatial data. In this work, we use the FakeNewsNet data repository~\cite{shu2020fakenewsnet} named to mitigate the lack of quality datasets. To our knowledge, FakeNewsNet is the only dataset that contains features about the content, social context and spatiotemporal information as shown in its statistics in Table~\ref{tab:FakeNewsNet}.

Our work helps this dataset to have more diversity with more social context features. First, the Twitter usernames of the most often referenced named entities were manually added to the dataset. Secondly, posts from named entities mentioned in the input post and the legitimacy score of the source enrich the FakeNewsNet dataset with more social context features.

We chose to use a subpart of the data due to the number of experiments to launch and the execution time of each experiment. Table \ref{tab:ExpData} shows the distribution of the data for our experimentation. This system also has a manually created list of official Twitter accounts of the most mentioned named entities of the used dataset.

\begin{table}[h]
\caption{Experimental-data distribution}\label{tab:ExpData}
\centering
\begin{tabular}{|l|l|l|l|}
\hline
\textbf{Features} & \textbf{\begin{tabular}[c]{@{}l@{}}Training\\  subset\end{tabular}} & \textbf{\begin{tabular}[c]{@{}l@{}}Validation \\ subset\end{tabular}} & \textbf{\begin{tabular}[c]{@{}l@{}}Test \\ subset\end{tabular}} \\ \hline
Content of articles & 120 & 15 & 15 \\ \hline
Articles author’s name & 112 & 13 & 14 \\ \hline
\begin{tabular}[c]{@{}l@{}}Article published \\ date and time\end{tabular} & 120 & 15 & 15 \\ \hline
Content of tweet & 12000 & 1500 & 1500 \\ \hline
\begin{tabular}[c]{@{}l@{}}Tweet author’s \\ name (username)\end{tabular} & 9812 & 878 & 941 \\ \hline
\begin{tabular}[c]{@{}l@{}}Tweet published\\  date and time\end{tabular} & 11868 & 1282 & 1330 \\ \hline
\end{tabular}
\end{table}

\subsubsection{Data Processing}
Due to the time complexity of training and fine-tuning our models, we decided to limit the number of features of the FakeNewsNet dataset. The entire dataset was only used to compare the performance of ExFake with other approaches. Only articles and tweets published or posted during the year 2016 or directly related to this specific year were considered. For instance, we consider tweets posted at the beginning of the year 2017 or the end of the year 2015 but related to articles published in 2016 (due to the huge propagation of fake news during the 2016 US presidential election). We also kept all tweets, including information about the author (i.e., user), the label, as well as the publication date and time. These tweets are the input of ExFake. Table~\ref{tab:tDataset} shows the result of the feature transformation. The pre-processing of data is then performed as follows: all the texts are transformed to lowercase, then stop words, punctuation signs, emojis and symbols are removed. After that, the data is transformed into tokens. The final step was to stem our dataset.

\begin{table}[h]
\caption{Features of the transformed dataset}\label{tab:tDataset}
\centering
\begin{tabular}{|l|l|l|}
\hline
\textit{\textbf{Aspect}} & \textit{\textbf{Source}} & \textit{\textbf{Features}} \\ \hline
\multicolumn{1}{|c|}{\multirow{2}{*}{Content-based}} & News article & The text content of the article \\ \cline{2-3} 
\multicolumn{1}{|c|}{} & Post (tweet) & The text content of the tweet \\ \hline
\multirow{4}{*}{Context-based} & \multirow{2}{*}{News article} & Author’s name \\ \cline{3-3} 
 &  & Published date and time \\ \cline{2-3} 
 & \multirow{2}{*}{Post (tweet)} & Author’s name \\ \cline{3-3} 
 &  & Published date and time \\ \hline
\end{tabular}
\end{table}

\subsubsection{Results and Discussions}
With the four system modules defined in Section.\ref{sc:approach}, we used Sentence-BERT (SBERT) for text-similarity and for natural language inference (NLI), Spacy tool for named entity recognition (NER), and Twitter web scraping API for data retrieval. Particularly in ExFake, for the text-similarity task, we use the SBERT a pre-trained BERT network that uses Siamese and triplet network architectures to generate semantically relevant sentence embedding. We fine-tuned SBERT using the STS benchmark (Semantic Textual Similarity)~\cite{DCMDEAILGLS}, a benchmark on sentence similarity. Table~\ref{tab:SBERThyperparam} presents the used hyper-parameters for the training phase. For the NLI task, SBERT was fine-tuned using the SNLI dataset \cite{bowman2015large}. Table \ref{table:SBERT_tab_nli} presents the used hyper-parameters for the training phase.

\begin{table}[h]
\centering
\caption{SBERT training hyper-parameters for the text similarity task}\label{tab:SBERThyperparam}
\begin{tabular}{|l|l|}
\hline
\textit{\textbf{Hyper-Parameter}} & \textit{\textbf{Value}} \\ \hline
Epochs number & 4 \\ \hline
Optimizer & Adam optimizer \\ \hline
Learning rate & 2e-5 \\ \hline
Weight decay & 0.01 \\ \hline
Regression loss & Mean squared-error loss \\ \hline
Batch-size & 16 \\ \hline
Random seeds & 10 \\ \hline
\end{tabular}
\end{table}

\begin{table}[h]
\centering
\caption{SBERT training hyper-parameters for the NLI task}
\label{table:SBERT_tab_nli}
\begin{tabular}{|c|c|}
\hline
\textbf{Hyper-Parameter}        & \textbf{Value}          \\ \hline
Epochs number                    & 5  \\ 
\hline
Optimizer                       & Adam optimizer          \\ \hline
Learning rate                   & 1e-5                    \\ \hline
Weight decay                    & 0.01                    \\ \hline
Regression loss                 & Sparse categorical cross-entropy loss \\ \hline
Batch-size                      & 128                      \\ \hline
\end{tabular}
\end{table}

For the named entity recognition (NER) task we used the SpaCy tool to extract the named entities. SpaCy extracts different named entities and regroups similar types of entities into labels such as GPE (for countries, cities, states, etc.), PERSON (for named person or family), ORG (for organizations, companies, agencies, institutions, etc.), DATE (for dates), LOC (for locations) and MONEY (for Monetary values, including unit), etc. Our system only keeps three labels: ORG, PER, and GPE.

For the explainability task and to provide an explanation based on the NLI task we followed a similar idea as Lime (Local Interpretable Model-agnostic Explanations)~\cite{ribeiro2016should}. The objective is to find the most important words, which led to the prediction. We start by removing a word from the hypothesis sentence. Then, we observe how the prediction score varies. The higher the variation the more we are certain that the removed word is essential. {Once we have the list of important words to highlight, the explanation is finally provided.}

Table~\ref{table:benchmark} presents a summary of the performance comparison between ExFake and the other seven models on the basis of the performance evaluation metrics, macro-accuracy, macro-recall, macro-precision and macro-F1-score.

\begin{table}[h]
\caption{Best performance comparison for fake news detection on FakeNewsNet}
\centering
\label{table:benchmark}
\begin{tabular}{|l|llll|}
\hline
\multirow{2}{*}{\textbf{Model}} & \multicolumn{4}{c|}{\textbf{Metric}} \\ \cline{2-5} 
 & \multicolumn{1}{l|}{\textbf{\begin{tabular}[c]{@{}l@{}}Macro-\\ accuracy\end{tabular}}} & \multicolumn{1}{l|}{\textbf{\begin{tabular}[c]{@{}l@{}}Macro-\\ precision\end{tabular}}} & \multicolumn{1}{l|}{\textbf{\begin{tabular}[c]{@{}l@{}}Macro-\\ recall\end{tabular}}} & \textbf{\begin{tabular}[c]{@{}l@{}}Macro \\ F1-score\end{tabular}} \\ \hline
SVM & \multicolumn{1}{l|}{0.580} & \multicolumn{1}{l|}{0.611} & \multicolumn{1}{l|}{0.717} & 0.659 \\ \hline
Logistic regression & \multicolumn{1}{l|}{0.642} & \multicolumn{1}{l|}{0.757} & \multicolumn{1}{l|}{0.543} & 0.633 \\ \hline
Naive Bayes & \multicolumn{1}{l|}{0.617} & \multicolumn{1}{l|}{0.674} & \multicolumn{1}{l|}{0.630} & 0.651 \\ \hline
CNN & \multicolumn{1}{l|}{0.629} & \multicolumn{1}{l|}{0.807} & \multicolumn{1}{l|}{0.456} & 0.583 \\ \hline
\begin{tabular}[c]{@{}l@{}}Social Article \\ Fusion /S\end{tabular} & \multicolumn{1}{l|}{0.654} & \multicolumn{1}{l|}{0.600} & \multicolumn{1}{l|}{0.789} & 0.681 \\ \hline
\begin{tabular}[c]{@{}l@{}}Social Article \\ Fusion /A\end{tabular} & \multicolumn{1}{l|}{0.667} & \multicolumn{1}{l|}{0.667} & \multicolumn{1}{l|}{0.579} & 0.619 \\ \hline
Social Article Fusion & \multicolumn{1}{l|}{0.691} & \multicolumn{1}{l|}{0.638} & \multicolumn{1}{l|}{0.789} & 0.706 \\ \hline
\textbf{ExFake} & \multicolumn{1}{l|}{\textbf{0.808}} & \multicolumn{1}{l|}{\textbf{0.841}} & \multicolumn{1}{l|}{\textbf{0.871}} & \textbf{0.855} \\ \hline
\end{tabular}
\end{table}

The macro-accuracy of the baselines remains more or less close to 0.65 while ExFake could achieve 0.808. The macro-precision of the baselines ranges from 0.600 for the SAF/S to 0.807 for the CNN method. Our system slightly exceeded this score with a macro-precision of 0.841. Among the baseline approaches, the SAF got 0.789, which is the best macro-recall on par with the SAF/S. The CNN approach with a score of 0.456 has the worst macro-recall score. ExFake reaches a score of 0.871. The highest performance increase is on the macro F1-score where ExFake scored 0.855. The method that got the worst macro F1-score among the baselines is the CNN method with a score of 0.583. The best method between the baselines is the SAF, which got 0.706. ExFake outperforms all the state-of-the-art baselines of the FakeNewsNet benchmark on all the metrics.

The experiments to find out the impact of each module helped us to highlight the performance of two other aspects that we proposed for the fake news detection problem, which are the legitimacy score and the combination of the two NLP tasks. When used alone, the module that returns the legitimacy score based on the Bayesian average had an F1-score of 68\%. This score is better than five of the seven state-of-the-art baseline approaches. This behaviour is the same for the F1-score at the third time step for the modules that encapsulate the text similarity and NLI tasks.

ExFake's contribution to the fake news detection problem goes beyond the explanation provided by the system and the improvement of the state-of-the-art model results. The proposed model is the first to encapsulate the result of both text-similarity and natural language inference (NLI) tasks. Our experiments demonstrate that this encapsulation performs better than most state-of-the-art baselines. Additionally, our work enriches the FakeNewsNet dataset by incorporating a legitimacy score and data obtained from named entities identified within the input, thereby adding new social context elements to the dataset.

\section{Conclusion and Future Work}\label{sc:conc}
In this paper, we propose the ExFake system, an explainable fake news detection approach. We focus on the content of the post, as well as on context-based auxiliary information and data coming from a trusted fact-checking organization and named entities. ExFake is composed of four modules working together to return the percentage of confidence of an input tweet and provide interpretable explanations to OSN users. The explanations aim to improve their ability to spot fake news. Experimental results on a real-world fake news dataset demonstrate the effectiveness of the proposed framework and the importance of the combination of text similarity, natural language inference and data processing tasks for fake news prediction.
Future work aims to explore multimodal data, use different datasets, and add more fact-checker websites and other types of trusted sources. We intend to  conduct a real user survey to ensure that ExFake achieves its goal of motivating OSN users to adopt certain behaviours. We also plan to push our experimentation further for the Ex-Decision module in order to find out the minimum number of tweets from a user/source necessary to have a legitimacy score that best represents the "fake news history" of a user/source. Finally, we seek to relativize this score according to the topics in order to have an overall legitimacy score and also a sub-legitimacy score, which will be based on particular topics. 

\bibliographystyle{IEEEtran}
\bibliography{references}

\end{document}